\documentclass[letterpaper, 10 pt, conference]{ieeeconf}

\IEEEoverridecommandlockouts                              

\usepackage{amsmath,amssymb,amsfonts}
\usepackage{multirow}
\usepackage{graphicx}
\usepackage{rotating}
\usepackage{multicol}
\usepackage{amsmath}
\usepackage{float}
\usepackage{array}
\usepackage{caption}
\usepackage{subcaption}
\usepackage{float}
\usepackage{xcolor}
\usepackage{multirow}
\usepackage{graphicx}
\usepackage{color, soul, colortbl}

\usepackage{colortbl} 
\usepackage{hyperref}
\usepackage[utf8]{inputenc}

\begin{document}
\title{\LARGE Vision-Based Autonomous Navigation for Unmanned Surface Vessel in Extreme Marine Conditions }
\author{Muhayyuddin Ahmed$^{1}$, Ahsan Baidar Bakht$^{1}$, Taimur Hassan$^{1,2}$, Waseem Akram$^{1}$, Ahmed Humais$^{1}$\\Lakmal Seneviratne$^{1}$, Shaoming He$^{3}$, Defu Lin$^{3}$, and Irfan Hussain$^{1,*}$
\thanks{$^{1}$ Khalifa University Center for Autonomous Robotic Systems (KUCARS), Khalifa University, United Arab Emirates.}%
 \thanks{$^{2}$ Department of Electrical, Computer and Biomedical Engineering, Abu Dhabi University, UAE.}
  \thanks{$^{3}$ Beijing Institute of Technology, China.}
\thanks{$^{*}$ Corresponding Author, Email: irfan.hussain@ku.ac.ae}
}
\maketitle
\begin{abstract}
Visual perception is an important component for autonomous navigation of unmanned surface vessels (USV), particularly for the tasks related to autonomous inspection and tracking. These tasks involve vision-based navigation techniques to identify the target for navigation. Reduced visibility under extreme weather conditions in marine environments makes it difficult for vision-based approaches to work properly. To overcome these issues, this paper presents an autonomous vision-based navigation framework for tracking target objects in extreme marine conditions. The proposed framework consists of an integrated perception pipeline that uses a generative adversarial network (GAN) to remove noise and highlight the object features before passing them to the object detector (i.e., YOLOv5). The detected visual features are then used by the USV to track the target. The proposed framework has been thoroughly tested in simulation under extremely reduced visibility due to sandstorms and fog. The results are compared with state-of-the-art de-hazing methods across the benchmarked MBZIRC simulation dataset, on which the proposed scheme has outperformed the existing methods across various metrics.
\end{abstract}
\IEEEoverridecommandlockouts

%
\IEEEpeerreviewmaketitle

\section{Introduction}

Marine robotics has experienced significant growth in recent years due to the rapid growth of the maritime industry and the rising demand for ocean data~\cite{Robot_Survey}. Marine robots, particularly, unmanned surface vessels (USVs), are used to carry out a wide range of tasks, such as environmental monitoring and autonomous surveillance for maritime security. USVs are also used for rescue operations and for the irregular activity detection of
vessels around ports. To perform these tasks in the complex and dynamic marine environment, an advanced perception system that accurately captures environmental details is required. The robust and real time visual perception system improves the USV's autonomy and intelligence level~\cite{rev1}. 

Vision-based perception systems provide vital information about the environment for USV intelligent control. Visual servoing is a popular USV navigation technique used to implement vision-based intelligent control. This method uses vision-based feedback to steer the USV toward the specified target while avoiding collisions with other objects. Moreover, vision-based feedback is crucial for activities involving autonomous inspection and monitoring. The visual servoing-based approach often assumes that the camera can clearly view the target object/vessel in the environment.

The United Arab Emirates (UAE), experiences extreme
weather conditions, such as sandstorms in the summer and extreme fog in the winter. During sandstorms or extreme fog, the visibility in the air and water can be reduced significantly, making it difficult for autonomous systems such as USVs to perform navigation tasks effectively. Developing vision-based navigation/tracking approaches for extreme weather conditions with reduced visibility is still a challenging research problem. It is essential to develop advanced perception techniques that allow USV to navigate under such circumstances. The existing approaches for vision-based navigation, such as~\cite{rev1}~\cite{rev2}~\cite{rev3}, applied object detection algorithms for target detection. However, in harsh marine conditions such as extreme fog, high winds, and sandstorms along the coast, significantly affect the detector's
performance because the camera input to the detector is insufficient. Improving the camera input before passing it to the detector is necessary to enhance object detection accuracy. Recent advancements in computer vision have developed deep learning-based approaches for image enhancement and dehazing. These approaches are helpful to enhance image quality in such a way that the detector can easily perform the desired tasks. However, the use of these techniques in marine robotics for navigation is very limited.  

\begin{figure}
    \includegraphics[width=\columnwidth]{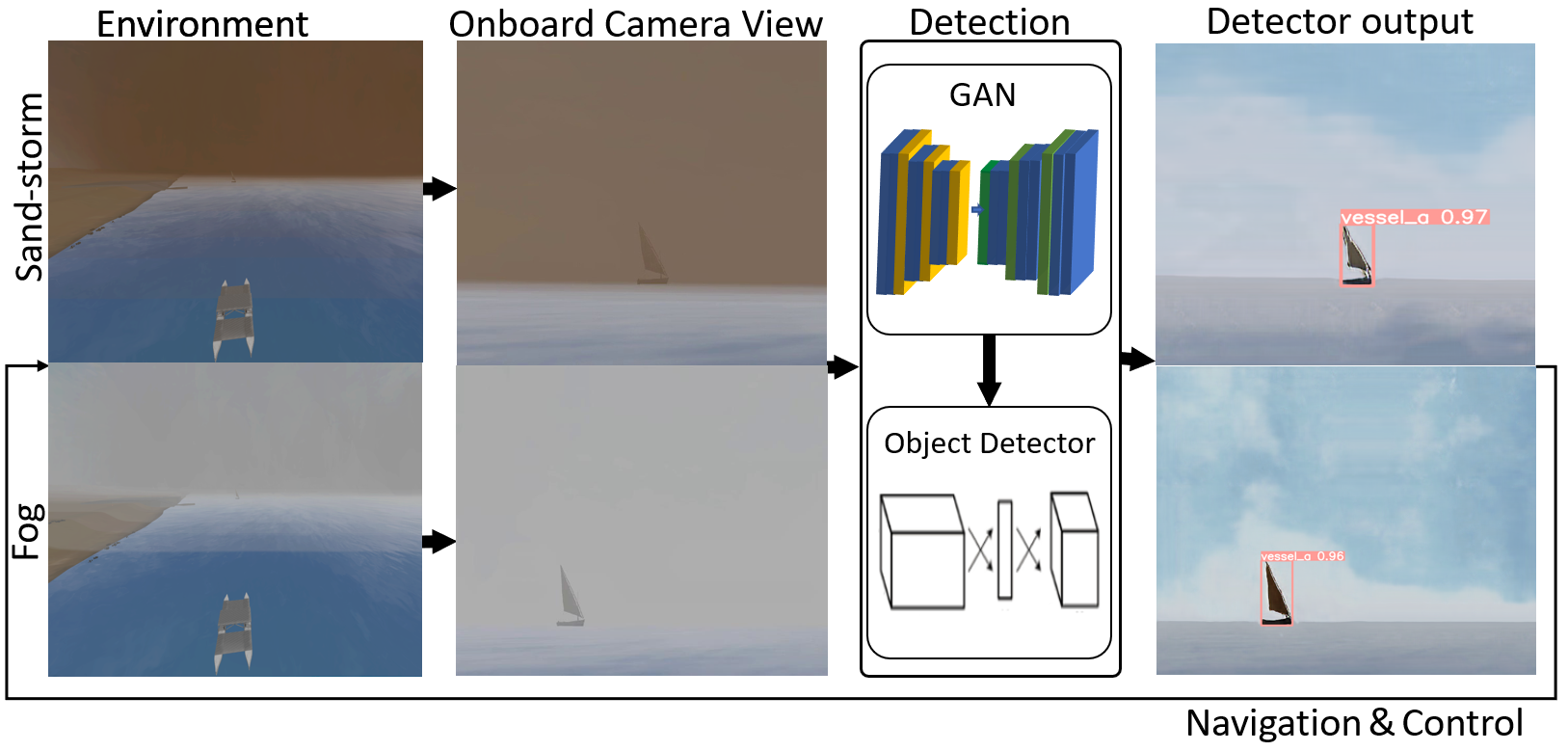}
    \caption{Overview of object detection and tracking in extreme conditions with reduced visibility. Onboard camera images are passed to the detection module, which performs image dehazing using GANs. The processed image passes to the detector, which detects the target object in the image and uses this information for USV navigation.}
    \label{fig:frontimg}
\end{figure}
This study contributes in this direction and proposes a visual servoing-based object tracking framework for extreme marine conditions such as sandstorms and extreme fog.  

\textit{Contributions:} The core contribution of this paper is the vision-based framework for target tracking in extreme conditions with highly reduced visibility in a coastal environment. The framework incorporates; 
 an integrated object detecting system that consists of: \textit{a)} generative adversarial networks (GANs) model for image dehazing, that minimizes the noise from the image and highlights the object features. \textit{b)} YOLOv5-based object detection system that takes an enhanced image as input to detect the target object, as depicted in Fig.~\ref{fig:frontimg}. 
In order to track the detected object using visual servoing, a PID-based feedback control system is implemented  that computes the control inputs for USV navigation to track the target object. It is important to note that the proposed framework is generalized; any image de-noising model and USV control technique can be used within the framework.

The rest of the paper is structured as follows, Sec.~\ref{sec:r-work} describes the related work, Sec.~\ref{sec:s-overview} provides an overview of the solution and discusses the framework. The proposed approach is discussed in Sec.~\ref{sec:p-approache}, which explains the dehazing method, target detection approach, and tracking method. The results are discussed in Sec.~\ref{sec:r-discussion}, and Sec.~\ref{sec:conclusion} concludes the work.

\section{Related Work}\label{sec:r-work}

This section describes the related work regarding object detection in marine environments. USVs autonomously perceive their environment using the onboard camera; the captured images are then used for detection. Object detection on the sea surface is challenging due to various factors, such as, weather conditions (fog and haze). In recent years, deep learning-based object detection techniques have emerged and are being used in various applications. Some of the commonly used detection techniques are the R-CNN series \cite{rr4}, covering Faster RCNN \cite{rr5}, Cascade R-CNN \cite{rr6}, Mask R-CNN \cite{rr7}, SSD \cite{rr8}, and YOLO \cite{rr9}. Despite extensive research on object detection techniques, their practical application in the marine environment is still very limited.

Object detection techniques are required for USV to successfully detect and track the desired object. For instance, \cite{rev1} used YOLOv4-based object detection to detect the objects with the USV onboard camera. In this work, a cross stage partial network (CSPNet) was deployed in a feature fusion network, which helps to reduce the computation burden. Additionally, objects with different resolutions were learned by tuning the weights during the training process. The YOLOv4 model was then integrated with the feature pyramid transformer (FPT) in order to obtain cross-space and cross-scale feature interactions. The proposed approach was trained and tested on a sea surface buoy object detection dataset collected in a marine environment. 

A fusion-based detection and tracking scheme for USV is proposed in~\cite{rev2}. In this work, high-resolution and deep semantic features are achieved by a cross-stage partial network. In addition, anchor and convolution layers are improved in the network model. The proposed model was tested on an image and video dataset. The results demonstrated better results with the fusion-based model compared with other detection and tracking algorithms. In \cite{rev3}, authors proposed an object detection scheme using radar-photoelectric approach. In this work, first-frame extraction is carried out, then region-of-interest (ROI) is used to predict the target object. The developed algorithm was tested in a marine environment with the help of a radar-guided target object. 

Despite the fact that numerous studies have been conducted to perform object detection and tracking in marine environments, the input data acquired from the environment is often blurry and noisy due to harsh conditions like extreme fog, winds, and sandstorms near the coast. In such conditions, detectors may not be able to detect the target object accurately because the camera input is inadequate. In such scenarios, improving the camera input before passing it to the detector is necessary to enhance object detection accuracy. To this end, deep learning-based image dehazing and enhancement techniques are helpful because they provide image quality improvement so that the detector can easily perform the desired tasks without mission failure.

Researchers, for example, have come up with a single image dehazing algorithm that uses deep convolutional neural networks. A fully convolutional network is used in the proposed method to turn a single hazy image into a clear image. The model learns to estimate the transmission map and atmospheric light in a data-driven manner~\cite{Dehaze_CNN}. In another approach, a multi-scale fusion network for single image dehazing is proposed. This method uses a multi-scale fusion strategy to estimate the transmission map and atmospheric light. The network consists of an encoder-decoder architecture with multi-scale feature fusion blocks to capture contextual information~\cite{Multi-Dehaze}. In~\cite{Pyramid_Dehaze}, a feature pyramid dehazing network is proposed for single-image dehazing. The network consists of an encoder-decoder architecture with FPN layers, which can estimate the transmission map and atmospheric light at multiple scales. The proposed method also introduces a texture preservation loss to preserve the high-frequency details of the dehazed image. Recently, the Transformer architecture has been extended to the computer vision domain and has shown promising results in image classification tasks. One of the early works that applied the Transformer to image dehazing is proposed by ~\cite{Transfromer_Dehaze}. The approach is a transformer-based network called DehazeTransformer, which can estimate the transmission map and atmospheric light simultaneously. The DehazeTransformer network takes the hazy image as input and provides the dehazed image as output. The self-attention mechanism in the Transformer allows the network to capture long-range dependencies in the input image and to generate high-quality dehazed images.

\begin{figure}
\begin{center}
    
    \includegraphics[width=0.9\columnwidth]{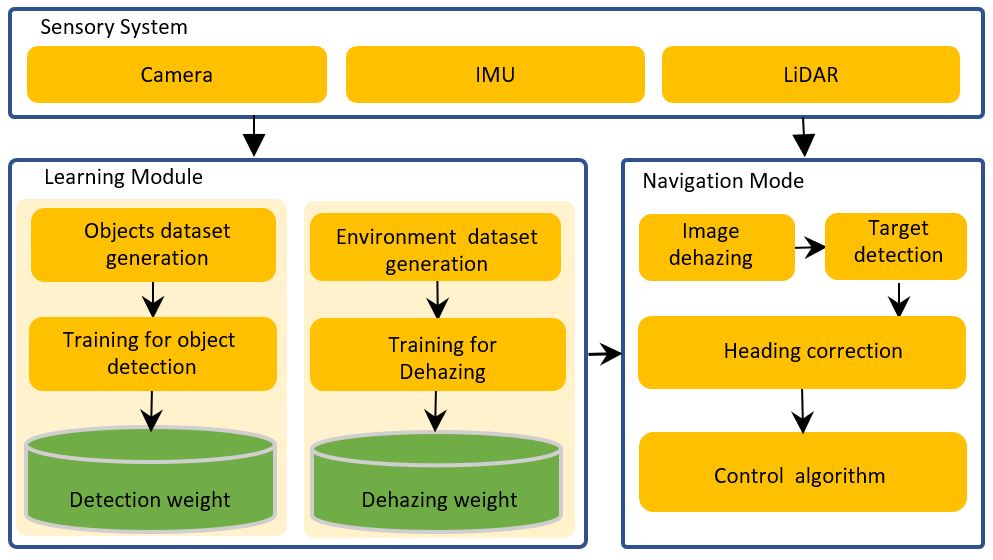}
    \caption{Framework for target tracking in extreme environment}
    \label{fig:framework}
    \vspace{-1.5em}
\end{center}

\end{figure}
\section{Solution Overview}\label{sec:s-overview}

The proposed framework consists of three main components: \textit{sensing system}, \textit{learning module}, and \textit{ navigation module}, as depicted in Fig.~\ref{fig:framework}. The sensing module consists of the USV's onboard sensory system, which includes an RGBD camera (used to perceive the environment), an IMU (used to get information about the USV's orientation), and a LiDAR (for collision avoidance).

The learning module consists of two main components: learning for object detection and image dehazing. The set of vessels that USV needs to track is used to prepare the dataset. The object detection model is trained over the collected dataset, and the trained model is saved for online processing. The vision pipeline for object detection is described in Sec.~\ref{sec:vision-for-detection}.  The training dataset for image dehazing is generated by navigating the USV in coastal environments under clear visibility, and this data will be used as ground truth. Then, another dataset is captured in low-visibility conditions caused by a sandstorm or fog. These two recorded datasets will be used to train the generative adversarial network. Sec.~\ref{sec:de-hazing} will explain the pipeline for dehazing.

Navigation mode will receive the image from the onboard camera and pass it to the image dehazing algorithm, which will reconstruct the image and pass it to the object detection module (that was trained during the learning phase). Once the object is detected, the heading correction module will compute the heading error and pass it to the control algorithm. The control algorithm will compute the appropriate controls to correct the USV orientation and start navigating towards the target; the detail of the navigation mode is covered in Sec. ~\ref{sec:visual-servoing}.

\section{Proposed Approach}\label{sec:p-approache}

\subsection{Vision for target detection }\label{sec:vision-for-detection}
Deep learning models are highly effective in object detection. In scientific literature, numerous object detectors are proposed, such as~\cite{efficientdet}~\cite{Detr}~\cite{fasterR}. In this study, we used YOLOv5, which is a very efficient and  commonly used model for object detection. The model is deployed on a custom dataset consisting of 6 different types of vessels ( Fig.~\ref{fig:dataset}).

In general, the YOLOv5 network model is composed of three components: the backbone, neck, and head. By combining gradient changes with the feature map, the CSPDarknet \cite{CSPnet}, which is used for feature extraction in the backbone, solves the issue of gradient information repetition during model training. It results in fewer parameters while increasing detection accuracy. 
FPN \cite{FPN} and PANet \cite{PANet} are fused to form the neck layer, which conducts multi-scale prediction across many layers and is in charge of feature fusion. It improves localization and semantic representation at various sizes. The convolutional batch normalization layers are finally concatenated in the final component, and subsequently, mask formation is performed by obtaining the pixel details.

The model uses a combined loss function in the prediction phase that combines bounding box regression, classification, and confidence and can be expressed as follows:

\begin{figure}
    \includegraphics[width=\columnwidth]{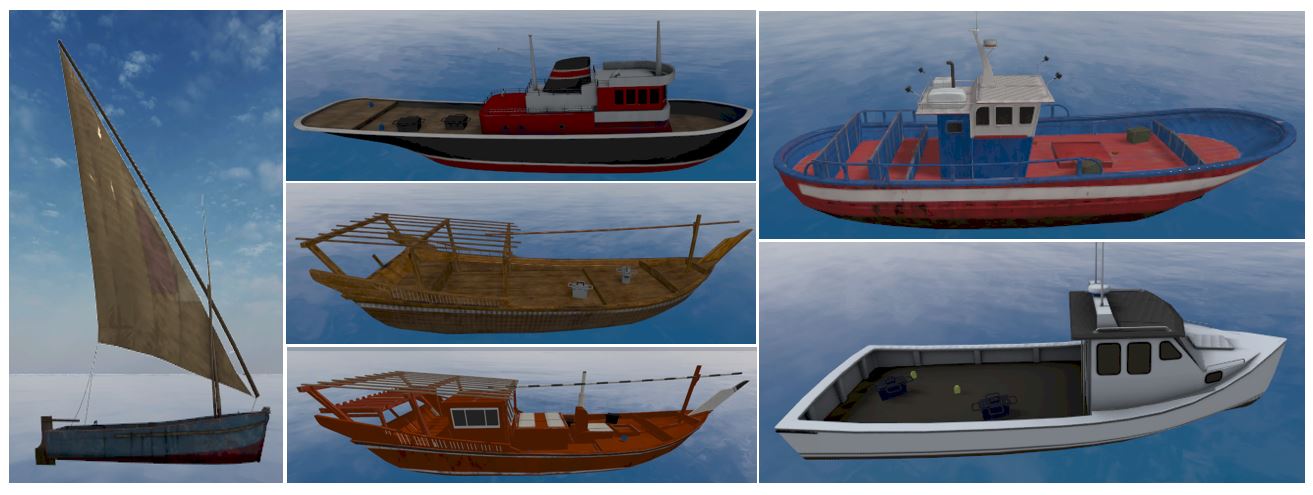}
    \caption{Dataset for vessel detection.}
    \label{fig:dataset}
        \vspace{-1.5em}
\end{figure}

\begin{equation}
    L=L_{clc}+L_{box}+L_{conf}
\end{equation}
where $L_{clc}$ is the classification error, $L_{box}$ the bounding box regression error, and $L_{conf}$ is the confidence error. 

The errors $L_{clc}$, $L_{box}$ , and $L_{conf}$  are computed as follows:

\begin{equation}
    L_{clc}=\sum_{i=0}^{k^2} l_i \sum_{c=1}^{C} E(\hat{p}_i(c),p_i(c))              
\end{equation}
where $l_i \in (1,0)$ for class objects as category c. The predicted and actual probabilities are represented by $\hat{p}_i$ and $p_i$.

\begin{equation}
    L_{box}=1-IoU(A^p,A^g)+\frac{A^c-A^p-A^g+I}{A^c}
\end{equation}
The bounding box, bounded by the predicted box is represented by $A^p$ and the actual box is represented by $A^g$. The desired and the actual areas are represented by $A^c$ and $I$, respectively. The ratio of the intersection between the estimated and actual areas within the image frame is denoted by $IoU$.

\begin{equation}
    L_{conf}= \sum_{i=0}^{k^2} \sum_{j=0}^{M} I_{i,j} E(\hat{C}_i, C_i) - \lambda_{no} \; \chi 
\end{equation}
\begin{equation}
     \chi = 
   \text{obj} \sum_{i=0}^{k^2} \sum_{j=0}^{M} I_{i,j} (1.I_{i,j}) E(\hat{C}_i, C_i)
\end{equation}
$k^2$ denotes the image partitioning into $k \times k$ grids, representing $M$ candidate anchors. $I_{i,j} \in (1, 0)$ distinguish positive or negative samples. $\hat{Ci}$ and $Ci$ are used to represent the confidence levels of the $i$th predicted bounding box and the actual bounding box, respectively. $\lambda_{no} \text{obj}$ shows the probability where the object $X$ does not exist. Additionally,  E(.) is used to denote a binary cross-entropy loss, which is defined as:
\begin{equation}
    E(\hat{X}_{i},X_{i})=\hat{X}_i ln (X_{i})+(1-\hat{X}_{i})ln(1-X_{i})
\end{equation}

This target object detection approach works with high detection accuracy in a clear environment. However, in cases of reduced visibility due to sand or fog, the performance of the detection algorithm is significantly reduced. 

\subsection{Vision pipeline for dehazing }\label{sec:de-hazing}
\begin{figure}
    \includegraphics[width=\columnwidth]{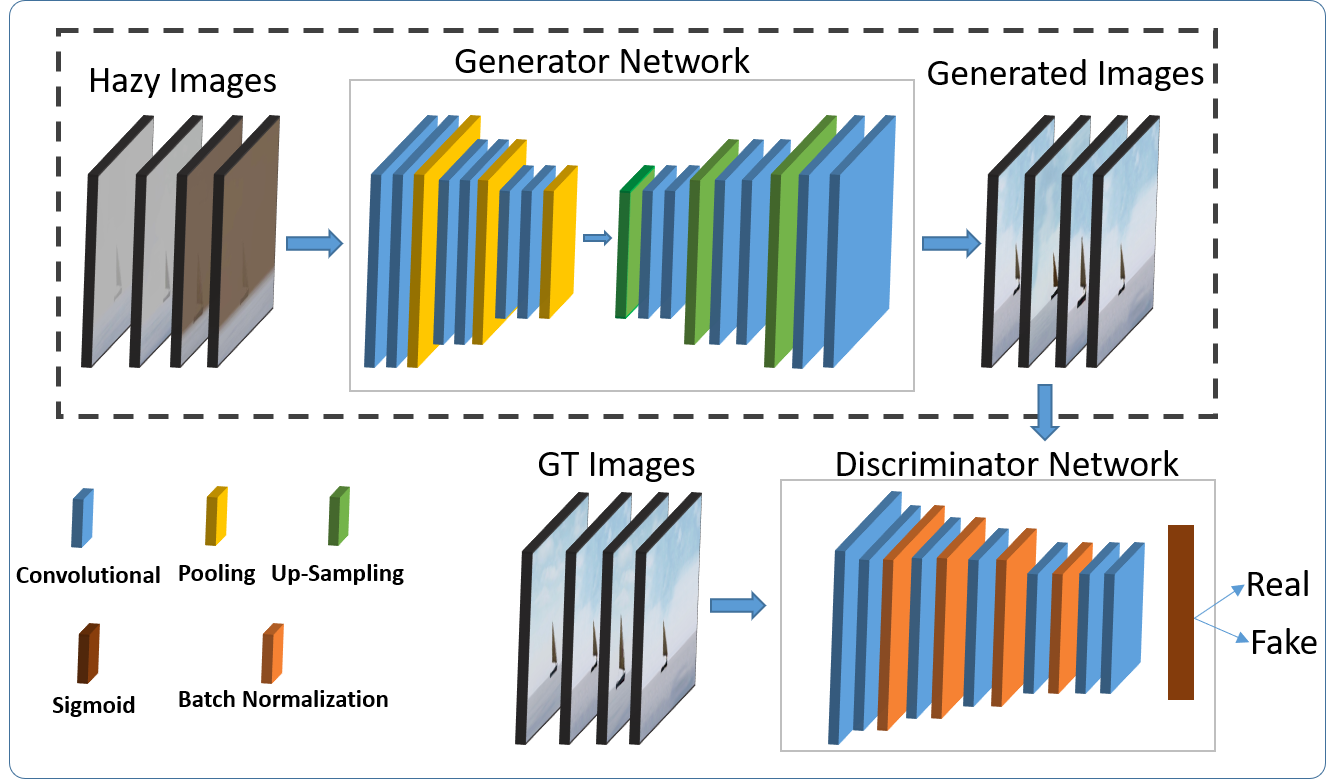}
    \caption{Architecture for image dehazing using generative adversarial networks (GANs): Input hazy images are fed to a generator network composed of an encoder-decoder architecture, which generates clear images. A discriminator network, consisting of sequential layers of convolution and batch normalization, is used to distinguish between real images and images generated by the generator network. This GAN-based approach is used for effective dehazing of images.}
    \label{fig:GAN_Dehazing}
    \vspace{-1.5em}

\end{figure}

The proposed method for image dehazing using GANs consists of a generator network and a discriminator network. The generator network takes a hazy image (i.e., a noisy image due to sandstorm or fog) as input and generates a clear image as output. The discriminator network takes a pair of hazy and clear images and distinguishes whether the clear image is a real or fake one, as shown in Fig.~\ref{fig:GAN_Dehazing}. During the training process, the generator network is trained to minimize the adversarial loss, which measures the difference between the generated and ground truth images according to the discriminator's feedback.

The generator network uses a pre-trained DenseNet-121~\cite{DenseNet} model for feature extraction and is modified to form an encoder-decoder structure with skip connections. The encoder part uses a series of transition blocks, residual blocks, and bottleneck block consisting of multiple convolutions and pooling layers, to refine the feature maps from the encoder part. The input hazy image is first downsampled through several convolutional layers in the encoder part of the network, gradually reducing the spatial dimensions of the input. This helps in capturing the essential features of the input image at multiple scales. The encoded feature map is then fed into the decoder part of the network, which upsamples the feature map through a series of transposed convolutional layers, gradually increasing the spatial dimensions of the output. This helps in reconstructing the dehazed image from the encoded features.

\begin{figure*}
    \includegraphics[width=\linewidth]{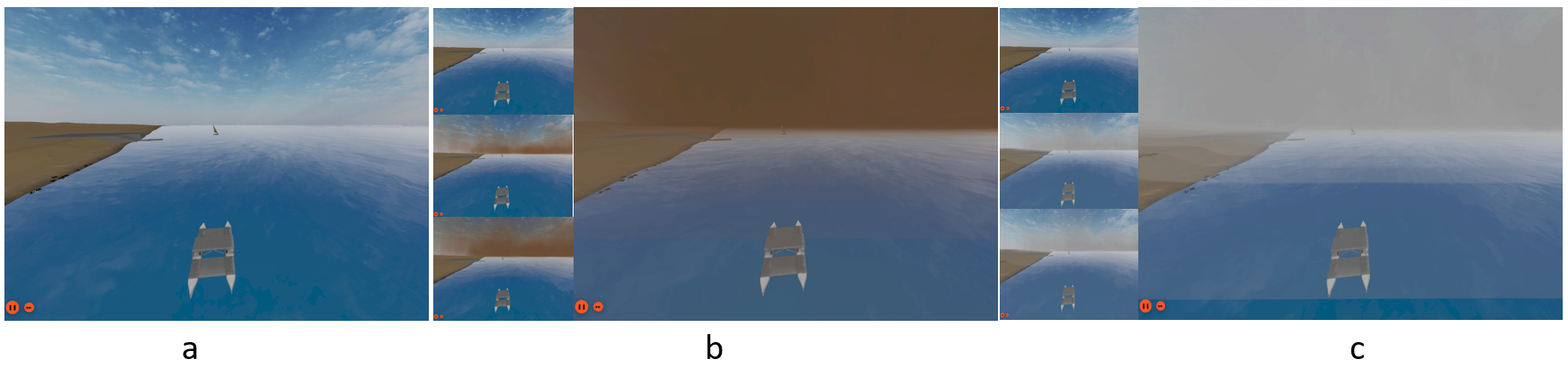}
    \caption {Marine environments under different weather conditions used to validate the proposed approach. Fig.-a represents the environment under clear weather conditions. Fig-b represents sandstorm, the small images shows different intensities of sandstorm. Fig-c represents the snapshot of extreme fog; the small figures show different intensity of the fog.}
    \label{fig:scene}
    \vspace{-1.5em}
\end{figure*}

\begin{figure}
    \includegraphics[width=\columnwidth]{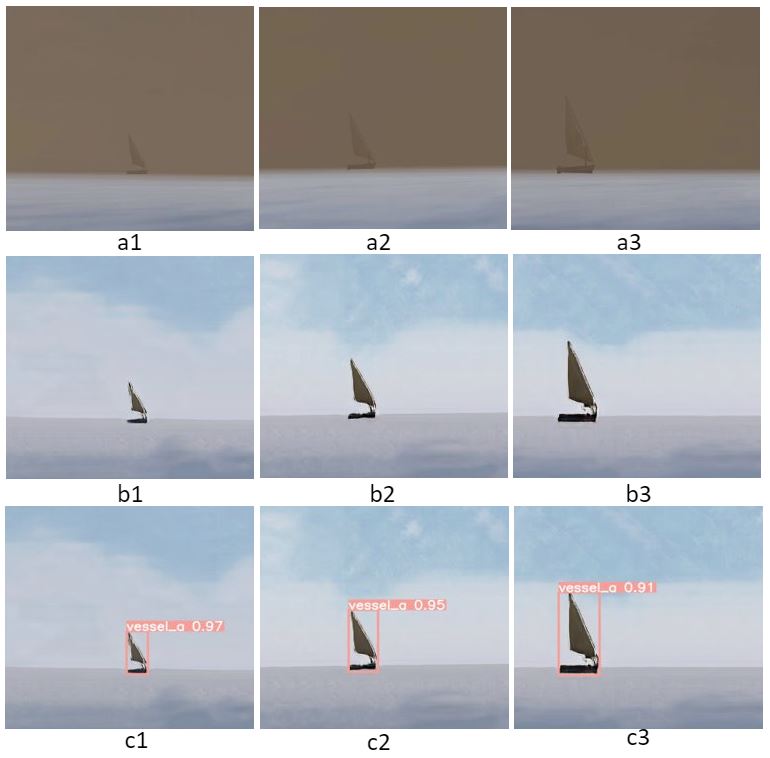}
    \caption {a1 to a3 represents the images captured by onboard camera from different distances during the sandstorm. b1 to b3 show the output of the GANs, and c1 to c3 show the output of the object detection module.}
    \label{fig:sequence}
      \vspace{-1.0em}

\end{figure}

\begin{figure}
    \includegraphics[width=\columnwidth]{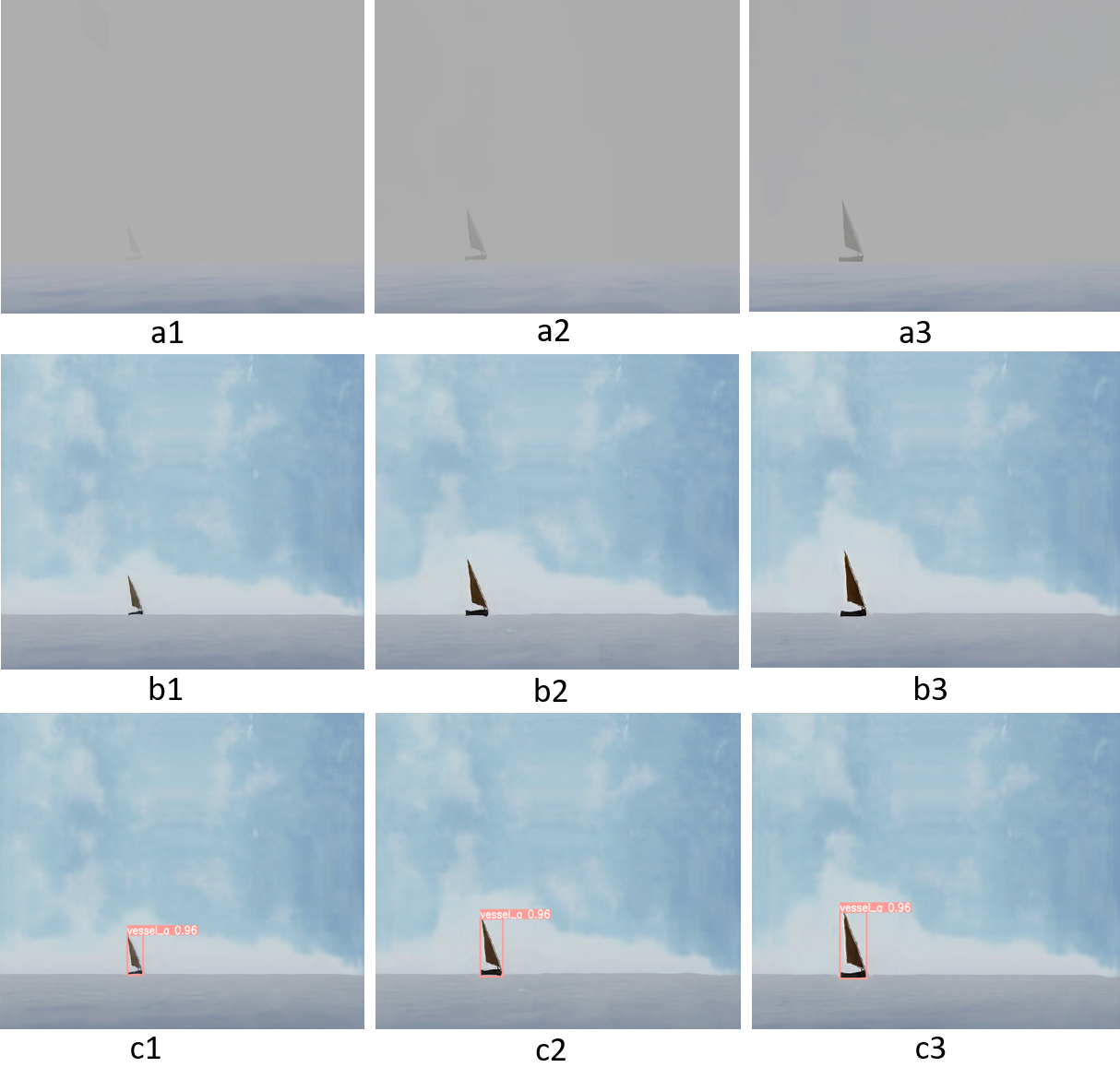}
    \caption {a1 to a3 represent the images captured by onboard camera from different distances during fog. b1 to b3 show the output of the GANs, and c1 to c3 show the output of the object detection module.}
    \label{fig:sequence1}
      \vspace{-1.5em}
\end{figure}

To train the generator model, we use multiple loss functions to calculate the total loss. Specifically, we use a combination of L1, Mean Squared Error (MSE), and Binary Cross Entropy (BCE) loss functions. 

\begin{equation}
L = L_{1} + L_{\mathrm{MSE}} + L_{\mathrm{BCE}}
\end{equation}

The $L_{1}$ loss denotes the absolute difference between the predicted output and the ground truth, while the $L_{MSE}$ represents the mean squared difference between the predicted output and the ground truth. $L_{BCE}$ is the difference between the predicted and ground truth binary labels, where the predicted labels are probabilities of the image being clear or hazy. The $L_{1}$, $L_{MSE}$, and $L_{BCE}$ can be defined mathematically as follows:
\begin{equation}
    L_1(y,\hat{y}) = \frac{1}{n} \sum_{i=1}^{n} |y_i - \hat{y_i}|
\end{equation}

\begin{equation}
    L_{MSE}(y, \hat{y}) = \frac{1}{n} \sum_{i=1}^{n} (y_i - \hat{y_i})^2
\end{equation}

\begin{equation}
    L_{BCE}(y, \hat{y}) = - \frac{1}{n} \sum_{i=1}^{n} \left[y_i \log \hat{y_i} + (1 - y_i) \log (1 - \hat{y_i})\right]
\end{equation}

where, $y$ represents the ground truth label, $\hat{y}$ denotes the predicted label, and $n$ is the number of images in the training dataset. The combination of these loss functions ensures that the generator model learns to produce images that are both visually similar to the ground truth images and have similar pixel values.

The discriminator model consists of a series of convolutional layers followed by batch normalization and LeakyReLU activation functions. The output of the last convolutional layer is then flattened and passed through a fully connected layer with a single output unit, this layer is followed by a sigmoid activation function to produce the final output of the discriminator. This output represents the probability, with values closer to 1 indicating a real image, and values closer to 0 indicating a fake image.

To create a dataset for training and testing GANs for the purpose of dehazing, a five-minute video was recorded in a simulation environment showing USV movement towards a target vessel in extreme conditions (sandstorm and fog) separately. The video was recorded using the USV onboard camera at a frame rate of 30 frames per second (fps). From the recorded video, 2 frames per second were extracted, resulting in a total of 600 images for each environment. The dataset was split equally into two sets, a training set and a testing set, with 300 images each. The training set of 300 images was used to train the GANs, and the testing set of 300 images was used to evaluate the performance of the trained model. A GAN was trained on a GPU machine for 100 epochs. The training system specifications include an Nvidia RTX A6000 GPU with 48GB of memory and an Intel Xeon Gold CPU 2.1GHzx80 processor.

After training the GAN model for dehazing, a network is made that can map hazy images to the ground truth that matches them. This generator network can be considered a model that has learned the underlying distribution of clear images given the hazy ones. The learned distribution captures the statistical regularities of clear images, allowing the generator to produce convincing clear images from hazy inputs. To apply this generator network for object detection tasks, we used it to preprocess the hazy images before passing them to the object detection model. This preprocessing step helps to remove the haze from the images, thus improving the detection performance of the object detection model.

\subsection{Visual servoing control}\label{sec:visual-servoing}

Image-based visual servoing (IBVS) is a way to control the movement of unmanned surface vessels by looking at images using on board camera. In IBVS, the camera captures images of the surrounding environment, and these images are processed to estimate the vessel's position and orientation relative to a desired trajectory or a target object. Based on this information, a control signal is generated to adjust the vessel's motion and bring it closer to the desired trajectory or object. IBVS can be used for a variety of tasks, including navigation, inspection, and object tracking. One of the main advantages of IBVS is that it does not require accurate measurements of the vessel's position or orientation since these can be estimated directly from the camera images. Additionally, IBVS can be used in environments where GPS or other external sensors may be unavailable or unreliable, such as in shallow or narrow waterways or in areas with poor satellite coverage.

To implement IBVS, a visual error function $e$ is defined that quantifies the difference between the desired visual features $\mathcal{F}_d$ and the observed visual features $\mathcal{F}_o$ in the image captured by the USV's onboard camera. 
Where, $\mathcal{F}_d$ represents that the detected target vessel should be in the center of the image. $\mathcal{F}_o$ describes the observed difference between the center of the image and the center of the bounding box of the target vessel. The error function $e$ is represented as follows: 

\begin{equation}
    e(t) = \mathcal{F}_d - \mathcal{F}_o
\end{equation}

The $e(t)$ is computed in terms of pixel error. The control input $u$ for the USV is computed using PID controller in the following way; 
\begin{equation}
    u(t) = K_\textrm{p} e(t) + K_\textrm{i} \int_{0}^{t} e(t)  \,dt + K_\textrm{d}\frac{de(t)}{dt}
\end{equation}

where $K_\textrm{p}$, $ K_\textrm{i}$, and $ K_\textrm{d}$ are all non-negative, and represent the coefficients for the proportional, integral, and derivative gain, respectively. The computed control input $u(t)$ applied to the USV for correcting the heading and navigation towards the target. The process of computing the error and applying the computed control input to the USV will continue until the USV reaches the desired proximity of the target vessel.  

\begin{figure}
    \includegraphics[width=\columnwidth]{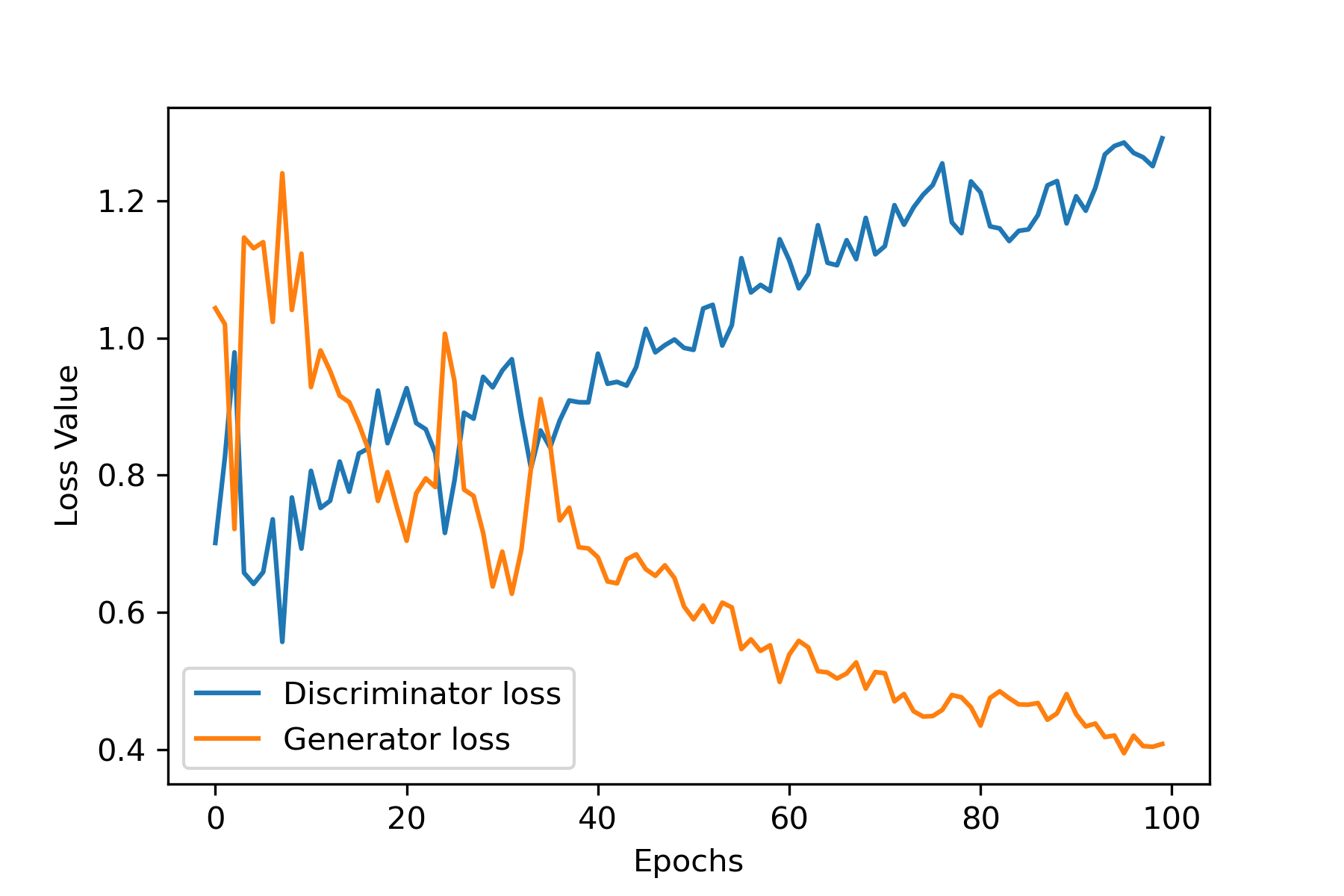}
    \caption{Training progress of the GAN-based architecture for image dehazing: The plot shows the variation of generator and discriminator network loss functions over 100 epochs of training.}
    \label{fig:Loss_Plot}
    \vspace{-1.5em}
\end{figure}

\section{Results and Discussions} \label{sec:r-discussion}

To validate the proposed approach, we used an open-source marine simulator (\url{https://github.com/osrf/mbzirc.git}) developed by Open Robotics for the MBZIRC Maritime Grand Challenge. It was developed in C++ and Python. It uses ROS2 Galactic for communication among modules. The hydrostatic and hydrodynamic features are modeled using Ignition Gazebo. This simulator also provides  the simulation of advanced features such as the generation of sandstorms and fog. Fig.~\ref{fig:scene} shows the different environments that are used to verify the proposed approach, Fig.~\ref{fig:scene}-a represents the clear environment with proper visibility, Fig.~\ref{fig:scene}-b shows the sandstorm, and Fig.~\ref{fig:scene}-c shows the reduced visibility due to fog. The implementation of the approach can be found in the GitHub repository\footnote{Source code: \url{https://github.com/Muhayyuddin/visual-servoing.git}}.

We navigate the USV towards the target vessel in a sandstorm or fog and apply a simple YOLOv5-based object detector; however, it was not able to detect the target vessel, particularly when the target was far from the USV. Under the same weather conditions, the proposed approach detected the object correctly. Fig~\ref{fig:sequence} and Fig.~\ref{fig:sequence1} show the sample images of sandstorm and fog.  Fig.~\ref{fig:sequence}-a1-3 and Fig.~\ref{fig:sequence1}-a1-3 show the sample images captured by the USV's onboard camera from different distances during sandstorm  and fog respectively. Fig.~\ref{fig:sequence}-b1-3 and Fig.~\ref{fig:sequence1}-b1-3 depict the output of the proposed dehazing module (GANs). The output of the GANs is forwarded to the YOLOv5 object detector, the detection results are shown in Fig.~\ref{fig:sequence}-c1-3 and Fig.~\ref{fig:sequence1}-c1-3.

Generator and discriminator loss curves over 100 epochs are shown in Fig.~\ref{fig:Loss_Plot}. It shows that the generator loss is decreasing over the 100 epochs, indicating that the model is learning to generate more realistic images as the training progresses. On the other hand, the discriminator loss is increasing, suggesting that the model is getting better at generating synthetic images, which makes it difficult to distinguish between real and generated images.

The results presented in Table~\ref{table1} indicate the performance of different backbone models within a proposed dehazing framework. It is worth noting that the choice of evaluation metrics is crucial when comparing the performance of different models. In our case, we use PSNR, which is used to evaluate the amount of signal distortion present in a dehazed image compared to a ground truth image; SSIM, which is used to assess the dehazing performance by measuring the contrast, luminance, and structure of both the ground truth and dehazed images; and MSE, which calculates the error between the dehazed image and the ground truth image. It can be seen that DenseNet-121 performs best in terms of PSNR, SSIM, and MSE with values of 24.99, 0.95, and 205.65, respectively. The second-best performance in terms of PSNR, SSIM, and MSE is achieved by ViT, with values of 24.55, 0.93, and 228.19, respectively. These results indicate that ViT and ResNet-50 are also viable options for dehazing tasks, but they are not as effective as DenseNet-121. 

The given Table~\ref{table2} shows a performance comparison of the proposed dehazing framework with other state-of-the art methods, including GW-FS, FDA, DehazeFormer, and FFA-Net. The comparative analysis is based on three commonly used image quality metrics: PSNR, SSIM, and MSE. The proposed dehazing block with DenseNet-121 as the backbone model outperforms all the compared state-of-the-art methods in terms of all three metrics. The proposed method achieves the highest PSNR of 24.99 dB, followed by DehazeFormer with a PSNR of 24.34 dB. In terms of SSIM, the proposed method again achieves the highest score of 0.95, followed by DehazeFormer with a score of 0.93. For MSE, the proposed method achieves the lowest value of 205.65, which is considerably lower than the other methods.

\begin{table}[t]
\caption {Effect of varying different backbone models within the proposed dehazing block. Bold indicates the best performance while the second-best performance is underlined.}    \label{table1}
\centering
\begin{tabular}{cccc}
\hline
BackBone     & PSNR  & SSIM & MSE    \\
\hline
DenseNet-121~\cite{DenseNet} & \textbf{24.99} & \textbf{0.95} & \textbf{205.65} \\
ResNet-50~\cite{ResNet}    & 24.08 & 0.92 & 253.84 \\
ViT~\cite{ViT}          & \underline{24.55} & \underline{0.93} & \underline{228.19} \\
MobileNet-v2~\cite{MobileNet} & 23.64 & 0.89 & 281.06 \\
VGG-16~\cite{Vgg-16}       & 23.92 & 0.86 & 304.82 \\
\hline
\end{tabular}

\end{table}

\begin{table}[t]
\centering
\caption{Performance evaluation of the proposed dehazing block with state-of-the-art dehazing schemes in terms of PSNR, SSIM, and MSE. Bold indicates the best performance, while the second-best performance is underlined.} \label{table2}
\begin{tabular}{cccc}
\hline
Framework     & PSNR  & SSIM & MSE    \\
\hline
Proposed & \textbf{24.99} & \textbf{0.95} & \textbf{205.65} \\
GW-FS~\cite{GW_FS}   & 23.90 & 0.90 & 264.73 \\
FDA~\cite{FDA}         & 23.08 & 0.84 & 319.51 \\
DehazeFormer~\cite{Transfromer_Dehaze} & \underline{24.34} & \underline{0.93} & \underline{238.92} \\
FFA-Net~\cite{FFA-NET}       & 23.41 & 0.87 & 296.08 \\
\hline
\end{tabular}
\vspace{-1.5em}
\end{table}

\section{Conclusion and Future Work}\label{sec:conclusion}
This paper presents a vision-based autonomous navigation framework for USV in extreme marine conditions such as sandstorm and fog near the coast. The proposed framework applies an integrated perception pipeline that uses a generative adversarial network to remove the noise and highlight the object features in the input image, captured by the USV's onboard camera. The enhanced image  is then passed to the YOLOv5 model to accurately detect the target object (vessel). The detection output is used by the visual servoing-based controller for tracking the target. The proposed scheme is compared with the state-of-the-art methods for dehazing on the benchmarked MBZIRC simulation dataset, where it significantly outperformed them across various metrics. 
In the future, the proposed approach will be tested in a real-world scenario, in which we will also enhance the USV control to incorporate extreme conditions that will include strong ocean currents and waves.  


\section*{Acknowledgement}
\noindent This work is supported by the Khalifa University of Science and Technology under Award No. MBZIRC-8434000194, CIRA-2021-085, FSU-2021-019, RC1-2018-KUCARS.

\bibliographystyle{ieeetr}
\bibliography{references}

\smallskip

\end{document}